\documentclass[10pt,twocolumn,letterpaper]{article}
\usepackage{cvpr}
\usepackage{times}
\usepackage{epsfig}
\usepackage{graphicx}
\usepackage{amsmath}
\usepackage{amssymb}

\usepackage{ctable}
\usepackage{multirow}
\usepackage{stfloats}
\usepackage{authblk}
\usepackage{array}
\newcolumntype{R}{>{\centering\arraybackslash}m{2.2cm}}
\newcolumntype{L}{>{\centering\arraybackslash}m{3cm}}
\newcolumntype{M}{>{\centering\arraybackslash}m{2.9cm}}
\newcolumntype{Q}{>{\centering\arraybackslash}m{3.5cm}}
\newcolumntype{V}{>{\centering\arraybackslash}m{1.62cm}}
\usepackage[ruled]{algorithm2e}
\usepackage[pagebackref=true,breaklinks=true,letterpaper=true,colorlinks,bookmarks=false]{hyperref}

\cvprfinalcopy 

\def\cvprPaperID{1222} 

\ifcvprfinal\fi
\begin{document}
\title{Efficient Image Set Classification \\using Linear Regression based Image Reconstruction}
\author[$\dag$]{S.A.A. Shah\thanks{The first two authors contributed equally to this work}}
\author[$\dag$]{U. Nadeem$^{*}$}
\author[$\dag$]{M. Bennamoun}
\author[$\ddagger$]{F. Sohel}
\author[$\dag$]{R. Togneri}
\affil[$\dag$]{The University of Western Australia}
\affil[$\ddagger$]{Murdoch University}
\affil[$\dag$]{\tt \small \{afaq.shah@, uzair.nadeem@research., mohammed.bennamoun@, roberto.togneri@\} uwa.edu.au}
\affil[$\ddagger$]{\tt \small f.sohel@murdoch.edu.au}

\maketitle

\begin{abstract}
We propose a novel image set classification technique using linear regression models. Downsampled gallery image sets are interpreted as subspaces of a high dimensional space to avoid the computationally expensive training step. We estimate regression models for each test image using the class specific gallery subspaces. Images of the test set are then reconstructed using the regression models. Based on the minimum reconstruction error between the reconstructed and the original images, a weighted voting strategy is used to classify the test set. We performed extensive evaluation on the benchmark UCSD/Honda, CMU Mobo and YouTube Celebrity datasets for face classification, and ETH-80 dataset for object classification. The results demonstrate that by using only a small amount of training data, our technique achieved competitive classification accuracy and superior computational speed compared with the state-of-the-art methods.
\end{abstract}

\section{Introduction}
Image set classification is defined as the problem of recognition from multiple images \cite{kim2007discriminative}. In image set classification, the gallery or training set consists of one or more image sets for each class and each image-set contains multiple images of the same class \cite{kim2007discriminative}. The test set also contains a number of images of the same subject which are then matched with the training image sets by computing some similarity measure to find the identity of the test subject.
\begin{figure}[t]
\begin{center}
   \includegraphics[width=1\linewidth]{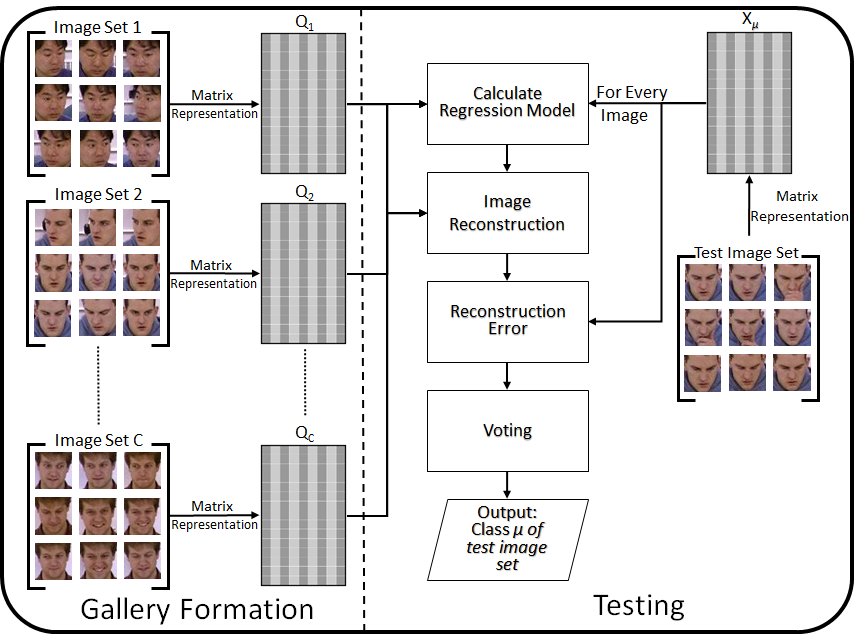}
\end{center}
   \caption{A block diagram of the proposed technique.}
\label{fig:blockDig}
\end{figure}
Compared with traditional single image based recognition, image set classification offers several advantages. For instance, image sets can effectively handle a wide variety of appearance variations within images including: viewpoint changes, occlusions, non-rigid deformation, variations in illumination and different backgrounds. Because of these characteristics, image set classification has been applied in many applications in biometrics including surveillance, video based face recognition and person re-identification in a network of security cameras \cite{hayat2015deep}.

Several image set classification techniques have been proposed in the literature. A few of these techniques, known as parametric methods \cite{arandjelovic2005face}, model image sets with certain statistical distributions and then calculate the simlarity between those distributions. However, these methods require a strong statistical relationship between the test and the training image sets to achieve good performance. As opposed to these methods, non-parametric methods represent image sets as linear or nonlinear subspaces \cite{ortiz2013face}, \cite{wang2012covariance}, \cite{yang2013face},  \cite{zhu2013point}. These methods have shown promising results and are being actively researched.

In this paper, we propose a novel non-parametric approach for image set classification. The proposed technique is based on the concept of image reconstruction using Linear Regression Classification (LRC) \cite{naseem2010linear} and nearest subspace classification. LRC uses the concept that samples of an image category lie on a linear subspace \cite{basri2003lambertian}, \cite{belhumeur1997eigenfaces}. In our proposed technique, the gallery image set of each category forms a subspace in a high dimensional space while using the downsampled images of each gallery image set. At test time, each image in the test image set is represented as a linear combination of images in each gallery image set. A least squares based solution is used to estimate the regression model parameters for each image of the test image set. The estimated regression model is used to reconstruct the test image from the gallery subspace. The Euclidean distance between the actual test image and the reconstructed image is then used as the distance metric. Next, weighted voting is used where each image of the test image set casts a vote for each class in the gallery. Finally, the decision rules in favor of the class with the highest accumulated weight. Figure \ref{fig:blockDig} shows the block diagram of the proposed technique. The performance of the proposed technique has been tested on four popular image set classification datasets CMU Motion of Body (CMU MoBo) Dataset \cite{gross2001cmu}, YoutubeCelebrity (YTC) Dataset \cite{kim2008face} and UCSD/Honda Dataset \cite{lee2003video} for face recognition and ETH-80 dataset \cite{leibe2003analyzing} for object recognition. We provide comparison with 13 image set classification algorithms. The main contributions of this paper can be summarized as follows:
\begin{itemize}
  \item A novel extension of LRC for image set classification, which is capable of producing state of the art results under the challenges of low resolution and less training data. The technique does not require any training and can easily be generalized  across different datasets. 
  \item Since LRC uses least squares solution, any technique using LRC is prone to the problem of singular matrix or singularity. This occurs when the rank is less than the number of rows in the regressor, a condition known as rank deficient matrix. While this problem is mostly ignored in the previous works of image set classification, we present practical and efficient solutions to overcome the problem of a rank deficient matrix. The solution is not limited to our technique and can be generalized to any method using LRC and least squares solutions.
  \item The techniques performing operations on each image of the test set are usually very slow and unsuitable for real time applications. On the other hand our technique uses an efficient matrix implementation of LRC to achieve the fastest test time compared to other image set classification methods.
\end{itemize}

The rest of this paper is organized as follows. An overview of related work is presented in Section \ref{related_work}. Section \ref{technique} discusses the proposed technique. Experimental results and detailed evaluation of the proposed technique against state-of-the-art approaches are presented in Section \ref{experiments}. The comparison of computational time of the proposed technique with other methods is presented in Section \ref{timing_analysis}. The technique is compared with other latest image set classification methods in Section \ref{discussion} and concluded in Section \ref{conclusion}. 

\section{Related Work} \label{related_work}
Image set classification techniques can be categorized as parametric, non-parametric and deep learning based methods. The parametric methods \cite{arandjelovic2005face} use a statistical distribution model to approximate an image set and then uses KL-divergence to measure the similarity between the two distribution models. Such methods, however, fail to produce good results in case of a weak statistical relationship between the training and the test image sets.  

For non-parametric methods, several different metrics are used to determine the set to set similarity. Wang et al. \cite{wang2008manifold}, use the Euclidean distance between the sets' mean as the similarity metric. Cevikalp and Triggs \cite{cevikalp2010face} present two models to learn set samples. The set to set distance using an affine hull model is called Affine Hull Image Set Distance (AHISD) while that using convex hull model is termed as the Convex Hull Image Set Distance (CHISD). Hu et al. \cite{hu2012face} used the mean image of image set and affine hull model to calculate the Sparse Approximated Nearest Points (SANP) for image sets in order to determine the distance between the training image set and test image set.
Some non-parametric methods (e.g., \cite{harandi2011graph}, \cite{wang2009manifold}, \cite{wang2012covariance}, \cite{wang2008manifold}) use a point on a geometric surface to represent the complete image set. 
The image set can also be represented either by a combination of linear subspaces or on a complex non-linear manifold. For linear subspaces, the cosine of the smallest angle between any vector in one subspace and any other vector in the other subspace is commonly used as the similarity metric between image sets.  

Discriminant analysis is commonly used to represent image sets on the manifold surface e.g., Discriminative Canonical Correlations (DCC) \cite{kim2007discriminative}, Manifold Discriminant Analysis (MDA) \cite{wang2009manifold}, Graph Embedding Discriminant Analysis (GEDA) \cite{harandi2011graph} and Covariance Discriminative Learning (CDL) \cite{wang2012covariance}. Chen \cite{chen2014dual} considered training and test image sets as subspaces in a high dimensional space and used the last image of each image set along with the variations of compared image sets through linear regression classification to determine the distance between the two subspaces. Feng et al. \cite{feng2016pairwise} extended the work of \cite{chen2014dual} by using the mean image of image sets instead of the last image. Moreover they also utilized the information which maximizes the distance between the test set and unrelated training sets. However, for these methods, the dimension of the feature vectors should be much larger than the combined number of images in the gallery and the test sets. Due to this limitation, these methods only work for small test sets.  
Hayat et al. \cite{hayat2014learning}, \cite{hayat2015deep} proposed a deep learning based approach called the Adaptive Deep Network Template (ADNT). In their technique, a deep autoencoder is used to define class-specific models for training sets. The weights of an autoencoder are initialized with a Gaussian Restricted Boltzmann Machine (GRBM). For classification, each image of the test set is reconstructed using a learnt class-specific model and reconstruction error is used as a measure to identify the test image set. ADNT has been demonstrated to achieve better performance compared to well-known methods, but it relies on hand crafted LBP features and requires fine tuning of several parameters for good performance . Moreover, training the ADNT requires a large number of images and is computationally expensive. Our technique reconstructs images in the test image set using LRC from the gallery image matrix and is much faster than ADNT both at training and test times. The proposed technique does not have any constraints on the number of images in the test set. Moreover, our technique can produce state of the art results using lower resolution images and much fewer training data, compared to other techniques. 
\section{Proposed Technique} \label{technique}
Let $N$ be the number of gallery images in each unique class $C$ of the gallery set $K_c$. Each image is converted to grayscale and downsampled to a resolution of $a \times b$ to be represented as $k_{c}^{n} \in \mathbb{R}^{a\times b}$, where $c=1,2,3,...,C$ and $n=1,2,3,...,N$. Each gallery image is transformed through column concatenation to a vector  such that $k_{c}^{n} \in \mathbb{R}^{a\times b} \rightarrow q_{c}^{n} \in \mathbb{R}^{T\times 1}$, where $T = ab$. Based on the concept that a linear subspace is formed by patterns from the same class \cite{basri2003lambertian}, a class specific model $Q_{c}$ is constructed for each class $c$ by horizontally concatenating the image vectors of class $c$.
\begin{equation}
\label{TrainMat}
Q_c = [q_c^1 q_c^2 q_c^3 ... q_c^N]\in \mathbb{R}^{T\times N},\quad c=1,2,3,...,C 
\end{equation}

In this way, each class $c$ is represented by  a vector subspace $Q_c$ called the \textit{regressor} for class $c$. Each vector $q_c^n,\ n=1,2,3,...,N,\:$ of the regressor $Q_c$ spans a subspace of $\mathbb{R}^{T\times 1}$ . 

Let the problem be to classify the unknown class $\mu$ of a test image set $Y_\mu$ with $M$ number of images in one of the classes $c=1,2,3,...,C$. Similar to the gallery images, each image of the test image set is converted to grayscale and downsampled to the resolution of $a \times b$ to be represented as $y_{\mu}^{m} \in \mathbb{R}^{a\times b}$ where ${\mu}$ is the unknown class and $m=1,2,3,...,M$. Each downsampled image is transformed through column concatenation to a vector  such that $y_{\mu}^{m} \in \mathbb{R}^{a\times b} \rightarrow x_{\mu}^{m} \in \mathbb{R}^{T\times 1}, $ where $T = ab$. The image vectors $x_{\mu}^{m},m=1,2,3,...,M$ are concatenated horizontally to create the test matrix $X_{\mu}$ 
\begin{equation}
\label{TestMat} 
X_{\mu} = [x_{\mu}^1 x_{\mu}^2 x_{\mu}^3 ... x_{\mu}^M]\in \mathbb{R}^{T\times M},
\end{equation}
where $\mu$ is the unknown class. If $X_{\mu}$ belongs to the $c^{th}$ class then it should be possible to represent the image vectors of $X_{\mu}$ as a linear combination of the gallery images from the same class i.e.,
\begin{equation}
\label{EqtoSolv} 
x_{\mu}^m=Q_c\gamma_c^m,\ m=1,2,...,M,\ c=1,2,...,C
\end{equation}  
where $\gamma_c^m \in \mathbb{R}^{N\times 1}$ is a vector of parameters. For the unique solution of Equation (\ref{EqtoSolv}) to exist, the condition $T\geq N$ must hold. Given that the condition holds, $\gamma_c^m$ can be estimated for test image vector $x_{\mu}^m$ and regressor $Q_c$ by using the least squares method \cite{friedman2001elements}, \cite{ryan2008modern}, \cite{seber2012linear}:
\begin{equation}
\label{Solv1} 
\gamma_c^m =(Q_c'Q_c)^{-1}Q_c'x_{\mu}^m,\ m=1,2,...,M\ c=1,2,...,C
\end{equation}  
where $Q_c'$ is the transpose of $Q_c$. The image vector $x_{\mu}^m$ can be reconstructed for the class $c$ using the parameters vector $\gamma_c^m$ and the regressor $Q_c$:
\begin{equation}
\label{Solv2} 
\widehat{x}_c^m=Q_c\gamma_c^m,\ m=1,2,...,M\ c=1,2,...,C
\end{equation}
\begin{equation}
\label{Solv3} 
\widehat{x}_c^m=Q_c(Q_c'Q_c)^{-1}Q_c'x_{\mu}^m
\end{equation}
where $\widehat{x}_c^m$ is the reconstructed image vector for $x_{\mu}^m$ from the regressor $Q_c$. $\widehat{x}_c^m$  can also be interpreted as the projection of $x_{\mu}^m$ on the $c^{th}$ subspace.

Instead of solving Equation (\ref{Solv3}) individually for each image vector $x_{\mu}^m$, it can be formulated in the matrix form to efficiently utilize the computational power of modern computers:
\begin{equation}
\label{EqtoSolvMat} 
X_{\mu}=Q_c\Gamma_c,\ c=1,2,...,C
\end{equation} 
where $\Gamma_c \in \mathbb{R}^{N\times M}$ is a matrix of parameters. $\Gamma_c$ can be calculated by using the least square estimation.
\begin{equation}
\label{Solv4} 
\Gamma_c =(Q_c'Q_c)^{-1}Q_c'X_{\mu},\ c=1,2,...,C
\end{equation}
\begin{equation}
\label{Solv5} 
\widehat{X}_c=Q_c\Gamma_c,\ c=1,2,...,C
\end{equation}
\begin{equation}
\label{Solv6} 
\widehat{X}_c=Q_c(Q_c'Q_c)^{-1}Q_c'X_{\mu}
\end{equation} 
where $\widehat{X}_c \in \mathbb{R}^{T\times M}$ is the  matrix of reconstructed image vectors for $X_{\mu}$ from the regressor $Q_c$. The reconstruction error between each test image $x_{\mu}^m$ and the reconstructed image $\widehat{x}_c^m$ is calculated using the Euclidean distance:
\begin{equation}
\label{Solv7} 
d_c^m = \left \| x_{\mu}^m-\widehat{x}_c^m \right \|_2,\ c=1,2,...,C, \ m=1,2,...,M
\end{equation}  
We experimented with different voting strategies. Majority voting produced good results on ETH-80 dataset and UCSD/Honda dataset. Nearest Neighbour Classification including 1-NN, 5-NN and 11-NN produced good results on YouTube Celebrity Dataset, UCSD/Honda dataset and MoBo Dataset. However, we required a voting strategy which performs consistently across all datasets. We also experimented with different versions of weighted voting. In weighted voting each image $m$ of the test image set casts a vote $\theta _c^m$ for each class $c$ to determine the class of the test image set $X_{\mu}$. We experimented using the Euclidean distance, the inverse of the Euclidean distance and the square of inverse Euclidean distance as weights. However, the best performance was achieved when using the exponential of the Euclidean distance in weighted voting. Hence, the weight of vote $\theta _c^m$ of each image $m$ is defined by the following equation:
\begin{equation}
\label{Solv8} 
\theta_c^m = e^{-\alpha d_c^m},\ c=1,2,...,C, \ m=1,2,...,M
\end{equation} 
where $\alpha$ is a constant. The accumulated weight for each class $c$ from each image of test set is given by:
\begin{equation}
\label{Solv9} 
\Theta_c = \sum_{m=1}^M \theta_c^m,\ c=1,2,...,C
\end{equation}
The class $c$ which gets the maximum accumulated weight from all the images $x_{\mu}^m$ of the test image set $X_{\mu}$ is decided as the class of the test image set: 
\begin{equation}
\label{Solv10} 
\mu = arg\:\underset{c}{max} (\Theta_c)\ c=1,2,...,C
\end{equation}
Algorithm \ref{algo1} provides the proposed image set classification technique.

\subsection{The Problem of Singularity} \label{singularity}
As mentioned before, for Equation (\ref{EqtoSolv}) and Equation (\ref{EqtoSolvMat}) to be well conditioned, the total number of pixels $T=ab$ in downsampled gallery image vectors $q_c^n$ must be greater than or equal to the number of gallery images $N$ in each regressor $Q_c$ i.e., $T\geq N$. However, even if this condition holds, it is possible for regressor $Q_c$ to be singular as one or more of the rows of $Q_c$ may come out to be linearly dependent on other rows. In this case, regressor $Q_c$ is called rank deficient due to the fact that $r<T$, where $r$ is the rank of $Q_c$. Therefore, it is not possible to use Equation (\ref{Solv1}) and Equation (\ref{Solv4}) to calculate the parameters vector $\gamma_c$ or parameters matrix $\Gamma_c$. In this paper we present two solutions for this problem:
\subsubsection{Perturbation}
The singularity of the regressor $Q_c$ can be overcome by regularizing the regressor $Q_c$ by adding a small perturbation term \cite{wang2012covariance}. We empirically found that by adding a matrix $\varepsilon $ with uniform random values in the range $-0.5\leq \epsilon \leq +0.5$ removes the singularity of the regressor $Q_c$ i.e.,
\begin{equation}
\label{perturbation}
Q_c^*=Q_c+\varepsilon, \ \varepsilon \in \mathbb{R}^{T\times N} and\ \forall \epsilon \in \varepsilon ,\  -0.5\leq \epsilon \leq +0.5
\end{equation}
Note that Equation (\ref{perturbation}) is implemented before any preprocessing and the values in matrix $Q_c$ are in the range of $0$ to $255$. In this way, the maximum possible change in the value of any pixel is $0.5$. We observed that there was no deterioration in the classification accuracy when using this method.  
\begin{algorithm}[t]
	\SetKwInOut{Input}{Input}
    \SetKwInOut{Output}{Output}
\Input{Gallery image sets $K_c$, where $c=1,2,3,...C$. Test image set $Y_{\mu}$.}
\Output{Class $\mu$ of test image set $Y_{\mu}$.}
\textbf{Gallery Formation:}\\
\For {$c$ in $1\ to\ C$}
{
\For{$n$ in $1\ to\ N$} {$q_c^n\in \mathbb{R}^{T\times 1}=downsample\: images\: to\: a\times b$\\and$\: vectorize$, where $T=ab$
}
$Q_c \in \mathbb{R}^{T\times N}=[q_c^1 q_c^2 q_c^3 ... q_c^N]$
}
\textbf{Testing:}\\
\For {$m$ in $1\ to\ M$}
{$x_{\mu}^m\in \mathbb{R}^{T\times 1}=downsample\: images\: to\: a\times b$\\and$\: vectorize$, where $T=ab$}
$X_{\mu}\in \mathbb{R}^{T\times M}=[x_{\mu}^1 x_{\mu}^2 x_{\mu}^3 ... x_{\mu}^M]$\\
\For{$c$ in $1\ to\ C$} {
\For{$m$ in $1\ to\ M$}{
$\gamma_c^m =(Q_c'Q_c)^{-1}Q_c'x_{\mu}^m$\\
$\widehat{x}_c^m=Q_c\gamma_c^m$\\
$d_c^m=\sqrt{\sum^{T}((x_{\mu}^m-\widehat{x}_c^m)^2)}$\\
$\theta_c^m = e^{-\alpha d_c^m}$
}
$\Theta_c = \sum_{m=1}^M \theta_c^m$}
$\mu = arg\:\underset{c}{max} (\Theta_c)$
\caption{The Proposed Image Set Classification Technique}
\label{algo1}
\end{algorithm}
\subsubsection{Basic Solution using QR decomposition}
In our second solution, we overcome the problem of singularity by computing a basic solution for Equation (\ref{EqtoSolv}) or Equation (\ref{EqtoSolvMat}) using QR decomposition \cite{press2007numerical}, \cite{stoer2013introduction} of the regressor $Q_c$ with the condition that the number of non-zero components in the solution vector $\gamma_c \leq r$, where $r$ is the rank of the regressor $Q_c$. This method does not remove the singularity of the regressor $Q_c$, however, the results obtained with this method are accurate for the purpose of our image reconstruction technique.

\subsection{Fast Linear Image Reconstruction} \label{fast}
A substantial decrease in the processing time can be achieved when using Equations (\ref{EqtoSolvMat}), (\ref{Solv4}), (\ref{Solv5}) and (\ref{Solv6}) compared to the use of Equations (\ref{EqtoSolv}), (\ref{Solv1}), (\ref{Solv2}) and (\ref{Solv3}). The processing time can further be reduced by calculating the inverse matrix of the regressor $Q_c$ using the Moore-Penrose pseudoinverse \cite{penrose1956best}, \cite{stoer2013introduction} at the time of gallery formation. In this way, the calculations at test time reduce to two matrix operations (Algorithm \ref{algo2}). Let $\widetilde{Q}_c$ be the pseudoinverse of the regressor $Q_c$ calculated at the time of gallery formation, then Equation (\ref{EqtoSolvMat}) can be solved at test time as:
\begin{equation}
\label{pseudoinverse}
\Gamma_c=\widetilde{Q}_c X_{\mu}
\end{equation}
\begin{equation}
\label{SolvPinv} 
\widehat{X}_c=Q_c(\widetilde{Q}_c X_{\mu})
\end{equation} 
In numerical analysis theory, the least squares solution using pseudoinverse is numerically less precise than using QR decomposition. However, we did not observe any degradation in the accuracy when using the pseudoinverse. Nearly two times gain in computational efficiency was achieved by the fast linear image reconstruction for ETH-80 dataset (refer to Section \ref{timing_analysis}). The gain in computational efficiency is more substantial for larger datasets. 

\begin{table*}[t]
  \centering
    \begin{tabular}{|l|M|M|M|M|}
    \hline
    \textbf{Methods$\downarrow$ \textbackslash $\ $Datasets$\rightarrow$} & \textbf{MoBo} & \textbf{YTC} & \textbf{Honda} & \textbf{ETH-80} \\
    \hline
    \textbf{TIS \cite{yamaguchi1998face}} & 96.81 $\pm$ 1.97 & 50.21 $\pm$ 3.59 & 88.21 $\pm$ 3.86 & 75.50 $\pm$ 4.83 \\
    \textbf{DCC \cite{kim2007discriminative}} & 88.89 $\pm$ 2.45 & 51.42 $\pm$ 4.95 & 92.56 $\pm$ 2.25 & 91.75 $\pm$ 3.74 \\
    \textbf{MMD \cite{wang2008manifold}} & 92.50 $\pm$ 2.87 & 54.04 $\pm$ 3.69 & 92.05 $\pm$ 2.25 & 77.50 $\pm$ 5.00 \\
    \textbf{MDA \cite{wang2009manifold}} & 80.97 $\pm$ 12.28 & 55.11 $\pm$ 4.55 & 94.36 $\pm$ 3.38 & 77.25 $\pm$ 5.46 \\
    \textbf{AHISD \cite{cevikalp2010face}} & 92.92 $\pm$ 2.12 & 61.49 $\pm$ 5.63 & 91.28 $\pm$ 1.79 & 78.75 $\pm$ 5.30 \\
    \textbf{CHISD \cite{cevikalp2010face}} & 96.52 $\pm$ 1.18 & 60.42 $\pm$ 5.95 & 93.62 $\pm$ 1.63 & 79.53 $\pm$ 5.32 \\
    \textbf{GEDA \cite{harandi2011graph}} & 84.86 $\pm$ 3.24 & 52.48 $\pm$ 4.45 & 91.28 $\pm$ 5.82 & 79.50 $\pm$ 5.24 \\
    \textbf{SANP \cite{hu2012face}} & 97.64 $\pm$ 0.94 & 65.60 $\pm$ 5.57 & 95.13 $\pm$ 3.07 & 77.75 $\pm$ 7.31 \\
    \textbf{CDL \cite{wang2012covariance}} & 90.00 $\pm$ 4.38 & 56.38 $\pm$ 5.31 & 98.97 $\pm$ 1.32 & 77.75 $\pm$ 4.16 \\
    \textbf{RNP \cite{yang2013face}} & 96.11 $\pm$ 1.43 & 65.82  $\pm$ 5.39 & 95.90 $\pm$ 2.16 & 81.00 $\pm$ 3.16 \\
    \textbf{MSSRC \cite{ortiz2013face}} & 97.50  $\pm$ 0.88 & 59.36 $\pm$ 5.70 & 97.95 $\pm$ 2.65 & 90.50  $\pm$ 3.07 \\
    \textbf{SSDML \cite{zhu2013point}} & 95.14 $\pm$ 2.20 & 66.24 $\pm$ 5.21 & 86.41 $\pm$ 3.64 & 81.00 $\pm$ 6.58 \\
    \textbf{DLRC* \cite{chen2014dual}} & 91.60 $\pm$ 2.78 & 65.55 $\pm$ 5.16 & 92.31 & NA \\
    \textbf{PLRC* \cite{feng2016pairwise}} & 93.74 $\pm$ 4.3 & NA    & 89.74 & NA \\
    \textbf{ADNT \cite{hayat2015deep}} & \textbf{97.92 $\pm$ 0.73} & \textbf{71.35 $\pm$ 4.83} & \textbf{100.00 $\pm$ 0.0} & \textbf{98.12 $\pm$ 1.69} \\\hline
    \textbf{Ours} & \textbf{98.33 $\pm$ 1.27} & \textbf{66.45 $\pm$ 5.07} & \textbf{100.00 $\pm$ 0.0} & \textbf{94.75 $\pm$ 4.32} \\\hline
    \end{tabular}%
    \caption[Average classification accuracies and standard deviations on CMU MoBo (MoBo)\cite{gross2001cmu}, YouTube Celebrity (YTC) \cite{kim2008face}, UCSD/Honda (Honda)\cite{lee2003video} and ETH-80 datasets. Both ;algorithms for the proposed technique have the same accuracy.;NA indicates the results of the respective methods are not available. * Indicates the use of a different experimental protocol than the one used for the other ;methods. Average accuracies are shown for the sake of completion.]
    {\tabular[t]{@{}l@{}}Average classification accuracies and standard deviations on CMU MoBo (MoBo) \cite{gross2001cmu}, YouTube Celebrity (YTC) \cite{kim2008face},\\ UCSD/Honda (Honda) \cite{lee2003video} and ETH-80 \cite{leibe2003analyzing} datasets. Both algorithms for the proposed technique have the same accuracy.\\{\footnotesize * Indicates use of different experimental protocol than the one used for the other methods. Average accuracies are shown for the sake of} \\{\footnotesize completion. NA indicates the results of the respective methods are not available.}\endtabular}
  \label{Mobo}%
  \label{YTC}%
  \label{Honda}%
  \label{ETH}%
\end{table*}%
\section{Experiments and Analysis}\label{experiments}
Extensive experiments were carried out to demonstrate the performance of our technique. We evaluated our technique on three video databases, namely CMU Motion of Body Dataset (CMU MoBo) \cite{gross2001cmu}, Youtube Celebrity Dataset (YTC) \cite{kim2008face} and Honda/UCSD Dataset \cite{lee2003video} for face recognition. ETH-80 \cite{leibe2003analyzing} was used for the task of object recognition. We selected these datasets because they are commonly used to evaluate image set classification techniques. 

We compared our technique with several prominent image set classification methods. These techniques include Face Recognition using Temporal Image Sequence (TIS) \cite{yamaguchi1998face}, Discriminant Canonical Correlation Analysis (DCC) \cite{kim2007discriminative}, Manifold-Manifold Distance (MMD) \cite{wang2008manifold}, Manifold Discriminant Analysis (MDA) \cite{wang2009manifold}, the Linear version of the Affine Hullbased Image Set Distance (AHISD) \cite{cevikalp2010face}, the Convex Hullbased Image Set Distance (CHISD) \cite{cevikalp2010face}, Graph Embedding Discriminant Analysis (GEDA) \cite{harandi2011graph}, Sparse Approximated Nearest Points (SANP) \cite{hu2012face}, Covariance Discriminant Learning (CDL) \cite{wang2012covariance}, Regularized Nearest Points (RNP) \cite{yang2013face}, Mean Sequence Sparse Representation Classification (MSSRC) \cite{ortiz2013face}, Set to Set Distance Metric Learning (SSDML) \cite{zhu2013point} and Adaptive Deep Network Template (ADNT) \cite{hayat2015deep}. We also compared our results with the Dual Linear Regression based Classifier (DLRC) \cite{chen2014dual} and Pairwise Linear Regression Model (PLRC) \cite{feng2016pairwise}. We followed the experimental protocols of Hayat et al. \cite{hayat2015deep} since they had the best classification accuracy. For comparison, we referenced the recognition results of \cite{cevikalp2010face}, \cite{harandi2011graph}, \cite{hayat2015deep}, \cite{hu2012face}, \cite{kim2007discriminative}, \cite{ortiz2013face}, \cite{wang2009manifold}, \cite{wang2012covariance}, \cite{wang2008manifold}, \cite{yamaguchi1998face}, \cite{yang2013face} and \cite{zhu2013point} reported in \cite{hayat2015deep}. DLRC \cite{chen2014dual} and PLRC \cite{feng2016pairwise} follow slightly different protocol for experiments. For the sake of completion we have reported the average results of these techniques, wherever available, as reported in the respective papers.
\begin{algorithm}[t]
	\SetKwInOut{Input}{Input}
    \SetKwInOut{Output}{Output}
\Input{Gallery image sets $K_c$, where $c=1,2,3,...C$. Test image set $Y_{\mu}$.}
\Output{Class $\mu$ of test image set $Y_{\mu}$.}
\textbf{Gallery Formation:}\\
\For {$c$ in $1\ to\ C$}
{
\For{$n$ in $1\ to\ N$} {$q_c^n\in \mathbb{R}^{T\times 1}=downsample\: images\: to\: a\times b$\\and$\: vectorize$, where $T=ab$
}
$Q_c \in \mathbb{R}^{T\times N}=[q_c^1 q_c^2 q_c^3 ... q_c^N]$\\
$\widetilde{Q}_c=pseudoinverse(Q_c)$
}
\textbf{Testing:}\\
\For {$m$ in $1\ to\ M$}
{$x_{\mu}^m\in \mathbb{R}^{T\times 1}=downsample\: images\: to\: a\times b$\\and$\: vectorize$, where $T=ab$}
$X_{\mu}\in \mathbb{R}^{T\times M}=[x_{\mu}^1 x_{\mu}^2 x_{\mu}^3 ... x_{\mu}^M]$\\
\For{$c$ in $1\ to\ C$} {
$\Gamma_c=\widetilde{Q}_c X_{\mu}$\\
$\widehat{X}_c=Q_c\Gamma_c$\\
$D_c=\sqrt{\sum^{T}((X_{\mu}-\widehat{X}_c)^2)}$\\
$\Theta_c = \sum_{m=1}^M e^{-\alpha D_c}$\\
}
$\mu = arg\:\underset{c}{max} (\Theta_c)$\\
\caption{Algorithm for Fast image reconstruction and Classification}
\label{algo2}
\end{algorithm}
\subsection{CMU MoBo Dataset}
The CMU Motion of Body Database (CMU MoBo) \cite{gross2001cmu} contains videos of 25 individuals walking on a treadmill, captured from six different viewpoints. Only the videos from the front camera are used for the purpose of image set classification. All the subjects except the last one has four different videos following different walking patterns. The four walking patterns are slow walk, fast walk, inclined walk and holding a ball while walking. The original purpose of this database was to advance biometric research on human gait analysis \cite{gross2001cmu}. We used the video sequences of the first 24 individuals, as they contain all four walking patterns, which is similar to the previous works \cite{chen2014dual}, \cite{feng2016pairwise}, \cite{hayat2015deep}. The frames of each video were considered as an image set. Similar to \cite{cevikalp2010face}, \cite{hayat2015deep}, \cite{hu2012face} and \cite{wang2008manifold}, we randomly selected the video of one walking pattern of each individual as the gallery image set and the other three walking patterns were considered as the test set. As mentioned in Section \ref{technique}, the number of images should be less than or equal to the number of pixels in the downsampled images. In practice, the number of images should be considerably lower than the number of pixels. We randomly selected a small number of frames i.e., 50 from each gallery video. The face from each frame was automatically detected using the Viola and Jones face detection algorithm \cite{viola2004robust}. Similar to \cite{hayat2015deep}, the images were resampled to the resolution of $40 \times 40$ and converted to grayscale. Histogram equalization was applied to increase the contrast of images. 
Different to \cite{chen2014dual}, \cite{feng2016pairwise}, \cite{hayat2015deep}, we did not use any LBP features, and performed experiments on raw images. We used $\alpha = 0.2$ in Equation (\ref{Solv8}). The experiments were repeated for 10 times with different random selections for the gallery and the test sets. We also used different random selections of the gallery images in each round to make our testing environment more challenging. We achieved the best classification accuracy on MoBo dataset among all compared techniques. Table \ref{Mobo} provides the average accuracy of our technique along with a comparison with other methods.
\subsection{YouTube Celebrity Dataset}
The YoutubeCelebrity (YTC) Dataset \cite{kim2008face} contains 1910 video clips of 47 celebrities and politicians. This is the largest dataset used for image set classification. These noisy real world videos, downloaded from YouTube, have low resolution and are recorded at high compression rates. The Viola and Jones algorithm \cite{viola2004robust} failed to detect faces for a large number of frames. Therefore, similar to \cite{hayat2015deep}, the Incremental Learning Tracker \cite{ross2008incremental} was used to track the faces in video clips. To initialize the tracker we used the initialization parameters provided by the authors of \cite{kim2008face}\footnote{\url{http://seqam.rutgers.edu/site/index.php?option=com_content&view=article&id=64&Itemid=80}}. Although the cropped face region was not uniform across frames, we decided to use the automatically tracked faces without any refinement. As proposed in \cite{hayat2015deep}, \cite{hu2012face}, \cite{wang2009manifold}, \cite{wang2012covariance}, \cite{wang2008manifold}, five fold cross validation was used for experiments. The datset was divided into five folds while minimizing the overlap between the various folds. Each fold contains 423 video clips with 9 video clips per individual. Out of 9 video clips per individual, three videos were randomly selected as the gallery set while the remaining six were used as six separate test image sets. All the tracked face images were resampled to the resolution of $30\times 30$ and converted to grayscale, following the protocol of \cite{hayat2015deep}. Histogram equalization was applied to enhance the contrast of images. For the gallery image set we randomly selected 20 images from each of the three gallery videos per individual per fold. If any gallery video clip had less than 20 frames, all the images of that video were used for gallery formation. In this way each gallery set had a maximum of 60 images. We used $\alpha = 10.5$ in Equation (\ref{Solv8}). Our technique achieved the highest accuracy among all the parametric and non-parametric methods. Deep Learning based ADNT \cite{hayat2015deep} has a better classification accuracy, however, it should be noted that our method uses significantly less training data compared to \cite{hayat2015deep} and is much faster than \cite{hayat2015deep} (refer to Section \ref{timing_analysis}). Moreover our technique does not require any parameter fine tuning or training which makes it more suitable for real life applications. Table \ref{YTC} summarizes the average accuracies of the different techniques on YouTube Celebrity dataset.
 
\subsection{UCSD/Honda Dataset}
The UCSD/Honda Dataset \cite{lee2003video} consists of 59 videos of 20 individuals. The number of videos for each individual varies from one to five. The database was originally developed to provide a standard video database to evaluate face tracking and recognition algorithms \cite{lee2003video}. All the videos contain significant head rotations and pose variations. Moreover some of the video sequences also contain partial occlusions in some frames. We followed the same experimental protocol as \cite{hayat2015deep}, \cite{hu2012face}, \cite{lee2003video}, \cite{wang2009manifold} and \cite{wang2008manifold}. The face from each frame of videos was automatically detected using Viola and Jones face detection algorithm \cite{viola2004robust}. Similar to \cite{hayat2015deep}, the detected faces were downsampled to the resolution of $20\times 20$ and converted to grayscale. Histogram equalization was applied to increase the contrast of images. The images were standardized by subtracting the mean and dividing by the standard deviation. We randomly selected one video from each of the $20$ individuals as the gallery image set while the remaining 39 videos were used as the test image sets. In order to keep the number of gallery images considerably less than the number of pixels (refer to section \ref{technique}), we randomly selected a small number of frames i.e., 50 from each gallery video. We used $\alpha = 0.2$ in Equation (\ref{Solv8}). To improve the consistency in scores we repeated the experiment 10 times with a different random selection of gallery images, gallery image sets and test image sets. Our technique achieved a perfect classification accuracy while using a significantly less number of gallery images. Table \ref{Honda} summarized the average identification rates of our technique compared to other image set classification techniques.
\begin{table*}[htbp]
  \centering
  
    \begin{tabular}{|l|l|Q|Q|Q|}
    \specialrule{.15em}{.0em}{.0em} 
    \textbf{Dataset $\downarrow$} & \textbf{Method $\downarrow$} & \textbf{Resolution used by \cite{hayat2015deep}} & \textbf{$\mathbf{20\times20}$ Resolution} & \textbf{$\mathbf{15\times15}$ Resolution} \\\specialrule{.15em}{.0em}{.0em} 
    \multirow{2}[0]{*}{\textbf{MoBo}} & ADNT \cite{hayat2015deep} & 97.92 $\pm$ 0.73 & 91.81 $\pm$ 2.40 & 90.56 $\pm$ 3.13 \\\cline{2-5}
          & \textbf{Ours}  & \textbf{98.33} $\pm$ \textbf{1.27} & \textbf{98.75} $\pm$ \textbf{1.38} & \textbf{99.31} $\pm$ \textbf{1.18} \\\specialrule{.15em}{.0em}{.0em}
    \multirow{2}[0]{*}{\textbf{Honda}} & ADNT \cite{hayat2015deep} & 100.00 $\pm$ 0.00 & 100.00 $\pm$ 0.00 & 99.74 $\pm$ 0.81 \\\cline{2-5}
          & \textbf{Ours}  & 100.00 $\pm$ 0.00 & 100.00 $\pm$ 0.00 & \textbf{100.00} $\pm$ \textbf{0.00} \\\specialrule{.15em}{.0em}{.0em}
    \multirow{2}[0]{*}{\textbf{ETH-80}} & ADNT \cite{hayat2015deep} & \textbf{98.12} $\pm$ \textbf{1.69} & 88.75 $\pm$ 6.26 & 90.25 $\pm$ 4.63 \\\cline{2-5}
          & \textbf{Ours}  & 94.75 $\pm$ 4.32 & \textbf{95.50} $\pm$ \textbf{4.04} & \textbf{92.75} $\pm$ \textbf{6.39}* \\\specialrule{.15em}{.0em}{.0em}
    \end{tabular}%
    \caption[Average classification accuracies and standard deviations on low resolutions of our technique compared with ADNT\cite{hayat2015deep}. {\footnotesize * The slight decrease in performance}; {\footnotesize is due to the fact that number of gallery images is nearly equal to the number of pixels (refer to Section \ref{technique} for details).}]
    {\tabular[t]{@{}l@{}}Average classification accuracies and standard deviations on the low resolutions of our technique compared with ADNT \cite{hayat2015deep}. \\{\footnotesize * The slight decrease in performance is due to the nearly equal number of gallery images and the number of pixels (refer to Section \ref{technique} for details).}\endtabular}
  \label{lowresol}%
\end{table*}%
\subsection{ETH-80 Dataset}
The ETH-80 dataset \cite{leibe2003analyzing}\footnote{\url{http://www.d2.mpi-inf.mpg.de/Datasets/ETH80}} consists of eight object categories including fruits, animals and vehicles. The object categories are apples, pears, tomatoes, cows, dogs, horses, cars and cups. Each object category has eight different image sets. Each image set consists of 41 images of the object taken from different view angles.
The cropped images containing only the object without any border area were used. The images were resized to the resolution of $32\times 32$ to follow the protocol of \cite{hayat2015deep}. The images were converted to grayscale and were standardized by subtracting the mean and dividing by the standard deviation. Similar to \cite{hayat2015deep}, \cite{kim2007discriminative}, \cite{wang2009manifold} and \cite{wang2012covariance}, five image sets of each object category are randomly selected as the gallery set while the other five are considered to be independent test image sets.  We used $\alpha = 0.2$ in Equation (\ref{Solv8}). We repeated the experiments 10 times for different random selections of gallery and test sets. The performance of our technique is comparable to the state of the art deep learning technique \cite{hayat2015deep}. Table \ref{ETH} summarizes the results of our technique compared to other methods.
\subsection{Experiments at low resolution}
We carried out further experiments at lower resolutions to demonstrate the efficacy of our technique. We also compared the performance of our technique with ADNT \cite{hayat2015deep} using low resolution data. We kept all other experimental settings the same as in the previous sections. For \cite{hayat2015deep}, we used the implementation provided by the authors and kept all prameters settings the same as recommended in their paper. The experiments were repeated $10$ times with random selections of gallery and test sets. Table \ref{lowresol} shows the average classification accuracies which demonstrates the superior performance of our technique at low resolution. On CMU MoBo dataset \cite{gross2001cmu}, the performance improves at lower resolution. For UCSD/Honda dataset \cite{lee2003video}, the classification accuracy remains a perfect score with the change in resolution. On ETH-80 Dataset \cite{leibe2003analyzing}, we achieved the best performance at $20\times20$ resolution. This is due to the fact that at $15\times 15$ resolution, the number of gallery images is nearly equal to number of pixels (refer to Section \ref{technique} for details). At $15\times 15$ resolution, we are able to achieve an average classification accuracy of 95.25\% by reducing the number of gallery images. Overall, there is no significant change in the classification accuracy of our technique with the change in resolution. Compared to ADNT \cite{hayat2015deep}, our technique consistently achieved better performance using low resolution data. This shows that our technique is more suitable to applications where the data is of very low resolution e.g., CCTV surveillance.
\section{Computational Time Analysis}\label{timing_analysis}
The proposed technique achieves the fastest timing performance as compared to other techniques. Table \ref{timetable} shows the training time for various methods and the test time required to classify an image set on the ETH-80 dataset using a modern CPU with 8 GB RAM.  The proposed technique requires no training. 
Although the proposed technique reconstructs each image of the test image set from all the gallery image sets, but due to the efficient matrix representation (refer to section \ref{technique} and section \ref{fast}), the achieved timing efficiency is superior to the other methods. 

\section{Discussion}\label{discussion}
Our proposed technique as well as DLRC \cite{chen2014dual} and PLRC \cite{feng2016pairwise} use linear regression for classification. However, our technique is remarkably different from DLRC and PLRC. DLRC considers both training and test images sets as subspaces of a high dimensional space and uses the distance between the test and training image sets to determine the class of the test image set. To determine the distance between subspaces, DLRC uses the last image of each image set along with the variations between the training and test image sets \cite{feng2016pairwise} to solve a linear regression problem. PLRC is an extension of DLRC. In PLRC instead of the last image of each image set, the mean image is used along with the concept of related and unrelated subspaces \cite{feng2016pairwise}. Both DLRC and PLRC have the limitation that the combined number of images in the test and the training image sets should be much less than the number of features in the feature vectors. In addition to the gallery image sets, they also use test image sets as regressors, which render them prone to the problem of rank deficient matrix at test time. Moreover they use LBP features \cite{ojala2002multiresolution} for certain datasets. Therefore, the performance of these methods using raw images cannot be generalized. 

Our technique is quite different from DLRC and PLRC as we treat each image in the test image set independently and consider them as points in a high dimensional space. We reconstruct each test image from the gallery subspaces and use weighted voting with the Euclidean distances between the original and reconstructed test images. The use of weighted voting increases the robustness of our system to any noise and outliers in the test image set. Our technique does not impose any constraints on the number of images in the test set and can work with any number of test images. Moreover, we performed all of our experiments on raw images to demonstrate the generalizability of our technique. In contrast to DLRC and PLRC, once any singularity is removed in the regressor $Q_c$ at the time of the gallery formation (refer to section \ref{singularity}), our technique is immune to the problem of rank deficient matrix at test time. This is due to the fact that the test image set is not used as a regressor. Our technique can process whole image sets simultaneously and also has the capability to process one image at a time and update the class decision in real time which makes it suitable for live video surveillance (e.g., CCTV).
\begin{table}[htbp]
  \centering
    \begin{tabular}{|l|R|L|}
    \hline
    \textbf{Methods $\downarrow$} & \textbf{Total Training Time (seconds)} & \textbf{Test Time per image set (seconds)} \\
    \hline
    TIS \cite{yamaguchi1998face}   &  NR & 0.045 \\
    \hline
    DCC \cite{kim2007discriminative}   & 13.36 & 0.311 \\
    \hline
    MMD \cite{wang2008manifold}   & NR & 8.43 \\
    \hline
    MDA \cite{wang2009manifold}   & \textbf{1.22} & \textbf{0.005} \\
    \hline
    AHISD \cite{cevikalp2010face} & NR & 0.095 \\
    \hline
    CHISD \cite{cevikalp2010face} & NR & 0.213 \\
    \hline
    GEDA \cite{harandi2011graph}  & 2.7 & 0.068 \\
    \hline
    SANP \cite{hu2012face}  & NR & 105.7 \\
    \hline
    CDL \cite{wang2012covariance}   & 76.21 & 1.40 \\
    \hline
    RNP \cite{yang2013face}   & NR & 0.027 \\
    \hline
    MSSRC \cite{ortiz2013face} & NR  & 4.78 \\
    \hline
    SSDML \cite{zhu2013point} & 21.92 & 0.577 \\
    \hline
    ADNT \cite{hayat2015deep}   & \textbf{278.8} & \textbf{0.026} \\
    \hline
    \textbf{Ours} & \textbf{NR} & \textbf{0.0046} \\
    \hline
    \textbf{Ours (Fast)} & \textbf{NR} & \textbf{0.0028} \\
    \hline
    \end{tabular}%
    \caption{Computational Time Analysis on ETH-80 dataset. {\footnotesize NR shows that the method does not require training.}}
  \label{timetable}%
\end{table}%
The accuracy of the proposed technique is superior to all parametric and non-parametric methods. The deep learning technique ADNT \cite{hayat2015deep} has a better accuracy on the Youtube Celebrity dataset and the ETH-80 dataset. However, ADNT needs a lot of training data and relies on handcrafted LBP features. ADNT uses Restricted Boltzman Machine for parameter initialization and requires a lot of fine tuning. On the other hand, our technique uses only a fraction of the training data and achieves comparable results, using only the raw images. Moreover, our technique has produced superior results at lower resolutions, compared to ADNT, and is ten times faster than ADNT at test time. Our technique can also be easily generalized to new data. The capability to work with less training data and at low resolution deems our technique suitable for scenarios where only scarce training data is available and where fast decisions are required.

\section{Conclusion}\label{conclusion}
In this work, a novel image set classification technique is proposed. The proposed technique uses linear regression to reconstruct images of the test image set from gallery image sets and uses the accumulative weighted reconstruction error to decide for the class of the test image set. The technique requires less training data compared to other image set classification methods and can work effectively at very low resolution. Extensive experimental analysis has been presented on a number of popular and challenging datasets to demonstrate the superior performance of our technique. Through the efficient matrix implementation, the proposed technique achieves the fastest performance time. The technique can easily be scaled from processing one frame at a time (for live video acquisition) to processing all of the test data at once (for faster performance). All these factors make our technique ideal for image set classification applications. 

{\small
\bibliographystyle{ieee}
\bibliography{bibFinal}
}

\end{document}